\documentclass{article}
\pdfoutput=1
\usepackage[final]{nips_2016}
\usepackage{amsmath}
\usepackage{amsfonts}
\usepackage{graphicx}
\usepackage[utf8]{inputenc} 
\usepackage[T1]{fontenc}    
\usepackage{url}            
\usepackage{booktabs}       
\usepackage{amsfonts}       
\usepackage{nicefrac}       
\usepackage{microtype}      
\usepackage[font={footnotesize}]{caption}
\usepackage{enumitem}

\usepackage{rotating}
\usepackage{array}
\usepackage{authblk}
\usepackage{floatrow}
\usepackage{subcaption}
\newfloatcommand{capbtabbox}{table}[][\FBwidth]

\setlist{nosep}

\usepackage{color}

\title{Overcoming catastrophic forgetting in neural networks}
\author[a]{James Kirkpatrick}
\author[a]{Razvan Pascanu}
\author[a]{Neil Rabinowitz}
\author[a]{Joel Veness}
\author[a]{Guillaume Desjardins}
\author[a]{Andrei A. Rusu}
\author[a]{Kieran Milan}
\author[a]{John Quan}
\author[a]{Tiago Ramalho}
\author[a]{Agnieszka Grabska-Barwinska }
\author[a]{Demis Hassabis}
\author[b]{Claudia Clopath}
\author[a]{Dharshan Kumaran}
\author[a]{Raia Hadsell}
\affil[a]{DeepMind, London,  N1C 4AG, United Kingdom}
\affil[b]{Bioengineering department, Imperial College London, SW7 2AZ, London, United Kingdom}
\date{}
\begin{document}

\maketitle
\begin{abstract}
    The ability to learn tasks in a sequential fashion is crucial to the
    development of artificial intelligence. Neural networks are not, in general,
    capable of this and it has been widely thought that \emph{catastrophic
    forgetting}
    is an inevitable feature of
    connectionist models. We show that
    it is possible to overcome this limitation and train networks that can
    maintain expertise on tasks which they have not experienced for a
    long time. Our approach remembers
    old tasks by selectively slowing down learning on the weights important for 
    those tasks.
    We demonstrate our approach is scalable and effective by solving
    a set of classification tasks
    based on the MNIST hand written digit dataset and by learning several Atari
    2600 games sequentially.
\end{abstract}
\section{Introduction}
Achieving artificial general intelligence requires that
agents are able to learn and remember many different tasks
\citet{legg2007universal}.  This is particularly difficult in real-world
settings: the sequence of tasks may not be explicitly labelled,
tasks may switch unpredictably, and any individual task may not recur for long
time intervals. Critically, therefore, intelligent agents must demonstrate a
capacity for \emph{continual learning}: that is, the ability to learn consecutive
tasks without forgetting how to perform previously trained tasks.

Continual learning poses particular challenges for artificial neural networks
due to the tendency for knowledge of previously learnt task(s) (e.g.\ task A) to be
abruptly lost as information relevant to the current task (e.g.\ task B) is
incorporated. This phenomenon, termed \emph{catastrophic forgetting}
\citep{french1999catastrophic, mccloskey1989catastrophic, mcclelland1995,
ratcliff1990connectionist}, occurs specifically when the network is trained
sequentially on multiple tasks because the weights in the network that
are important for task A are changed to meet the
objectives of task B. 
Whilst recent advances in machine learning and in
particular deep neural networks have resulted in impressive gains in
performance across a variety of domains (e.g.\ \citep{krizhevsky2012imagenet,
lecun2015deep}), little progress has been made in achieving
continual learning. 
Current approaches have typically ensured that data from
all tasks are simultaneously available during training. By interleaving data
from multiple tasks during learning, forgetting does not occur because the weights
of the network can be jointly optimized for performance on all tasks.
In this regime---often referred to as the \emph{multitask learning}
paradigm---deep learning techniques have been used to train single agents that
can successfully play multiple Atari games \citep{rusu2015policy,
parisotto2015actor}.
If tasks are presented sequentially, multitask learning can
only be used if the data are recorded by an episodic memory system
and replayed to the network during training. This approach (often
called system-level consolidation \citep{mcclelland1995}), 
is impractical for learning large numbers of tasks, as 
in our setting it would require the amount of memories being stored
and replayed to be proportional to the number of tasks.
The lack of algorithms
to support continual learning thus remains a key barrier to the development of
artificial general intelligence.

In marked contrast to artificial neural networks, humans and other animals
appear to be able to learn in a continual fashion \citep{cichon2015branch}.
Recent evidence suggests that the mammalian brain may avoid catastrophic forgetting
by protecting previously-acquired knowledge in neocortical circuits
\citep{cichon2015branch, hayashi2015labelling, yang2009stably, yang2014sleep}.
When a mouse acquires a new skill, a proportion of excitatory synapses are
strengthened; this manifests as an increase in the volume of individual dendritic spines of
neurons \citep{yang2009stably}. Critically, these enlarged dendritic spines
persist despite the subsequent learning of other tasks, accounting for
retention of performance several months later \citep{yang2009stably}. When these
spines are selectively ``erased'', the corresponding skill is forgotten
\citep{hayashi2015labelling, cichon2015branch}. This provides causal evidence
that neural mechanisms supporting the protection of these strengthened synapses
are critical to retention of task performance.
Together, these experimental findings---together with 
neurobiological models \citep{fusi2005cascade, benna2015computational}---suggest
 that continual learning in the
mammalian neocortex relies on a process of task-specific synaptic
consolidation,  whereby knowledge about how to perform a previously acquired
task is durably encoded in a proportion of synapses that are rendered less
plastic and therefore stable over long timescales. 

In this work, we demonstrate that task-specific synaptic consolidation offers a
novel solution to the continual learning problem for artificial intelligence.
We develop an algorithm analogous to synaptic consolidation for artificial
neural networks, which we refer to as \emph{elastic weight consolidation} (EWC for short). This algorithm slows down
learning on certain weights based on how important they
are to previously seen tasks. We show how
EWC can be used in supervised learning and
reinforcement learning problems to train several tasks sequentially
without forgetting older ones, in
marked contrast to previous deep-learning techniques.

\section{Elastic weight consolidation}

In brains, synaptic consolidation enables continual learning
by reducing the plasticity of synapses that are vital to previously
learned tasks. We implement an algorithm that performs a similar
operation in artificial neural networks by constraining important parameters
to stay close to their old values.
In this section we explain why we expect to find a solution to a new task
in the neighbourhood of an older one, how we implement the constraint,
and finally how we determine which parameters are important.

\begin{figure}
    \begin{center}
        \includegraphics[width=.4\textwidth,trim=0cm 0mm 0cm 0cm,clip]{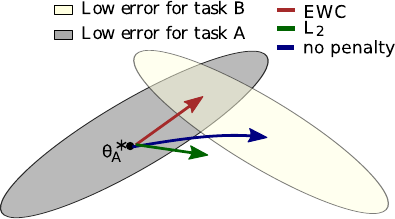}
    \end{center}
    \caption{\label{TR_depiction} elastic weight consolidation (EWC) ensures task A is remembered whilst training on task B.
    Training trajectories are illustrated in a schematic parameter space, with parameter regions leading to good performance
    on task A (gray) and on task B (cream).
    After learning the first task, the parameters are at $\theta^*_A$.
    If we take gradient steps according to task B alone (blue arrow), we will minimize the loss of task B but destroy what we have learnt for task A. On the other hand, if we constrain each weight with the same coefficient (green arrow) the restriction imposed is too severe and we can only remember task A at the expense of not learning task B. EWC, conversely, finds a solution for task B without incurring a significant loss on task A (red arrow) by explicitly computing how important weights are for task A. }
\end{figure}

A deep neural network consists of multiple layers
of linear projection followed by 
element-wise non-linearities.
Learning a task consists of adjusting the set of weights and biases $\theta$
of the linear projections, to optimize performance.
Many configurations of $\theta$ will result in the same performance
\citep{hechtnielsen:theory,Sussmann1992UniquenessOT};
this is relevant for EWC:
over-parameterization makes it likely that there is a
solution for task B, $\theta^*_B$, that is close to the previously found solution for task A,
$\theta^*_A$. While learning task B, EWC therefore
protects the performance in task A by constraining the parameters to
stay in a region of low error for task A centered around
$\theta^*_A$, as shown schematically
in Figure \ref{TR_depiction}.
This constraint is implemented as a quadratic penalty,
and can therefore be imagined as a spring anchoring the parameters to
the previous solution, hence the name elastic. Importantly,
the stiffness of this spring should not be the same for all
parameters; rather, it should be greater for those parameters
that matter most to the performance during task A.

In order to justify this choice of constraint and to define
which weights are most important for a task, it is useful to
consider neural network training from a probabilistic perspective.
From this point of view, optimizing the parameters is
tantamount to finding their most probable values given some data ${\cal D}$.
We can compute this conditional probability $p(\theta|{\cal D})$ from the
prior probability of the parameters $p(\theta)$ and the probability of the
data $p({\cal D}| \theta)$ by using Bayes' rule:
\begin{equation}
\label{Th1}
  \log p(\theta | {\cal D})  = \log p({\cal D}|\theta) + \log p(\theta) - \log p({\cal D})
\end{equation}
Note that the log probability of the data  given the parameters $\log p({\cal D}|\theta)$ is
simply the negative of the loss function for the problem at hand $-{\cal L}(\theta)$.
Assume that the data is
split into two independent parts, one defining task A (${\cal D}_A$) and the other task B (${\cal D}_B$).
Then, we can re-arrange equation \ref{Th1}:
\begin{equation}
  \label{Th2}
  \log p(\theta | {\cal D})  = \log p({\cal D}_B|\theta) +  \log p(\theta | {\cal D}_A) - \log p({\cal D}_B)
\end{equation}
Note that the left hand side is still describing the posterior probability of the parameters given the
\emph{entire} dataset, while the right hand side only depends on the loss function for task B
$\log p({\cal D}_B|\theta)$. All the information about task A must therefore have been absorbed
into the posterior distribution $p(\theta | {\cal D}_A)$. This posterior probability must contain
information about which parameters were important to task $A$ and is therefore the key to implementing
EWC.
The true posterior probability is intractable, so, following the work on the Laplace approximation by Mackay \citep{mackay1992practical}, we
approximate
the posterior as a Gaussian distribution with mean given by
the parameters $\theta^*_A$ and a diagonal precision given by the
diagonal of the Fisher information matrix $F$.
$F$ has three key properties \citep{pascanu2013revisiting}:
(a) it is equivalent to the second derivative of the loss
near a minimum, (b) it can be computed from first-order derivatives alone 
and is thus easy to calculate even for large models, and (c) it is guaranteed to be positive semi-definite.
Note that this approach is similar to expectation propagation where each subtask is seen as a factor of the posterior \citep{smola2003laplace}.
Given this approximation, the function ${\cal L}$ that we minimize in EWC is:
\begin{equation}
  \label{Th3}
  {\cal L}(\theta) = {\cal L}_B(\theta) + \sum_i \frac{\lambda}{2} F_i (\theta_i - \theta^*_{A,i})^2
\end{equation}
where $ {\cal L_B}(\theta) $ is the loss for task B only, $\lambda$ 
sets how important the old task is compared to the new one and $i$ labels each parameter.

When moving to a third task, task C, EWC will try to keep the network parameters
close to the learned parameters of both task A and B. This can be enforced
either with two separate penalties, or as one by noting that the sum of
two quadratic penalties is itself a quadratic penalty.

\subsection{EWC allows continual learning in a supervised learning context}

We start by addressing the problem of whether elastic weight consolidation
could allow deep neural networks to learn a set
of complex tasks without catastrophic forgetting.
In particular, we trained a fully connected multilayer
neural network on several supervised learning tasks in sequence.
Within each task, we trained the neural network in the traditional way, namely by shuffling
the data and processing it in small batches. After a fixed amount of training
on each task, however, we allowed no further training on that task's dataset.

We constructed the set of tasks from the problem of classifying hand written digits
from the MNIST \citep{lecun1998mnist} dataset, according to a scheme previously
used in the continual learning literature
\citep{srivastava2013compete, goodfellow2015empirical}. For each task, we generated
a fixed, random permutation by which the input pixels of all images would be
shuffled. Each task was thus of equal
difficulty to  the original MNIST problem, though a
different solution would be required for each. Detailed description of the settings 
used can be found in Appendix~\ref{app:mnist}.

\begin{figure*}
	\includegraphics[width=\textwidth]{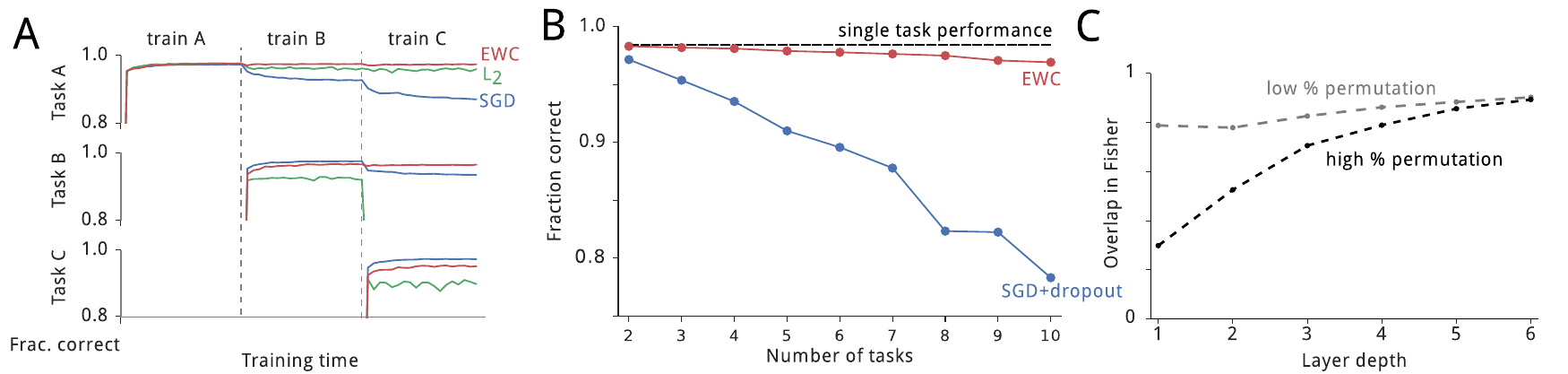}
    \caption{\label{MnistLearningCurves} Results on the permuted MNIST task. A: Training curves for three random permutations A, B and C using EWC(red), $L_2$ regularization (green) and plain SGD(blue). Note that only EWC is capable of mantaining a high performance on old tasks, while retaining the ability to learn new tasks. B: Average performance across all tasks using EWC (red) or SGD with dropout regularization (blue). { The dashed line shows the performance on a single task only.} C: Similarity between the Fisher information matrices as a function of network depth for two different amounts of permutation. Either a small square of 8x8 pixels in the middle of the image is permuted (grey) or a large square of 26x26 pixels is permuted (black). Note how the more different the tasks are, the smaller the overlap in Fisher information matrices in early layers.
}
\end{figure*}

Training on this sequence of tasks with plain stochastic
gradient descent (SGD) incurs catastrophic forgetting, as demonstrated in Figure
\ref{MnistLearningCurves}A. The blue curves show performance on
the testing sets of two different tasks. At the point at which the training
regime switches from training on the
first task (A) to training on the second (B), the performance for task B
falls rapidly, while for task A it climbs steeply. The forgetting of task
A compounds further with more training time, and the addition of subsequent tasks.
This problem cannot be countered by regularizing the network with a fixed quadratic
constraint for each weight (green curves, { L2 regularization}): here, the performance in task A
degrades much less severely, but task B
cannot be learned properly as the constraint protects all weights equally, leaving
little spare capacity for learning on B. However, when we use
EWC, and thus take into account how important each weight is to task
A, the network can learn task B well without forgetting task A (red curves).
This is exactly the expected behaviour described diagrammatically in Figure \ref{TR_depiction}.

Previous attempts to solve the continual learning problem for deep neural networks
have relied upon careful choice of network hyperparameters, together with other
standard regularization methods, in order to mitigate catastrophic
forgetting. However, on this task, they have only achieved reasonable results on
up to two random permutations \citep{srivastava2013compete,goodfellow2015empirical}.
Using a similar cross-validated hyperparameter search as \citep{goodfellow2015empirical},
we compared traditional dropout regularization to EWC.
We find that stochastic gradient descent with dropout regularization alone is limited, and that it does not scale to more
tasks (Figure \ref{MnistLearningCurves}B).  In contrast, EWC allows a large
number of tasks to be learned in sequence, with only modest growth in the error rates.

Given that EWC allows the network to effectively squeeze in more functionality into a
network with fixed capacity, we might ask whether it allocates completely separate
parts of the network for each task, or whether capacity is used in a more
efficient fashion by sharing representation. To assess this, we determined whether each task depends on the
same sets of weights, by measuring the overlap between pairs of tasks' respective
Fisher information matrices (see Appendix~\ref{app:fisher}). A small overlap means that the two tasks depend on different
sets of weights (i.e.\ EWC subdivides the network's weights for different tasks); a large overlap
indicates that weights are being used for both the two tasks (i.e.\ EWC enables sharing
of representations). Figure \ref{MnistLearningCurves}C shows the overlap  as a function of depth. As a simple control, when a network is trained on two tasks which are very similar to each other (two versions of MNIST where only a few pixels are permutated), the tasks depend on similar sets of weights throughout the whole network (grey curve). When then the two tasks are more dissimilar from each other, the network begins to allocate separate capacity (i.e.\ weights) for the two tasks (black line). Nevertheless, even for the large permutations, the {layers of the network closer to the output} are indeed being reused for both tasks. This reflects the fact that the permutations make the input domain very different, but the output domain (i.e.\ the class labels) is shared.

\subsection{EWC allows continual learning in a reinforcement learning context}

\begin{figure*}
	\begin{center}
		\includegraphics[width=\textwidth]{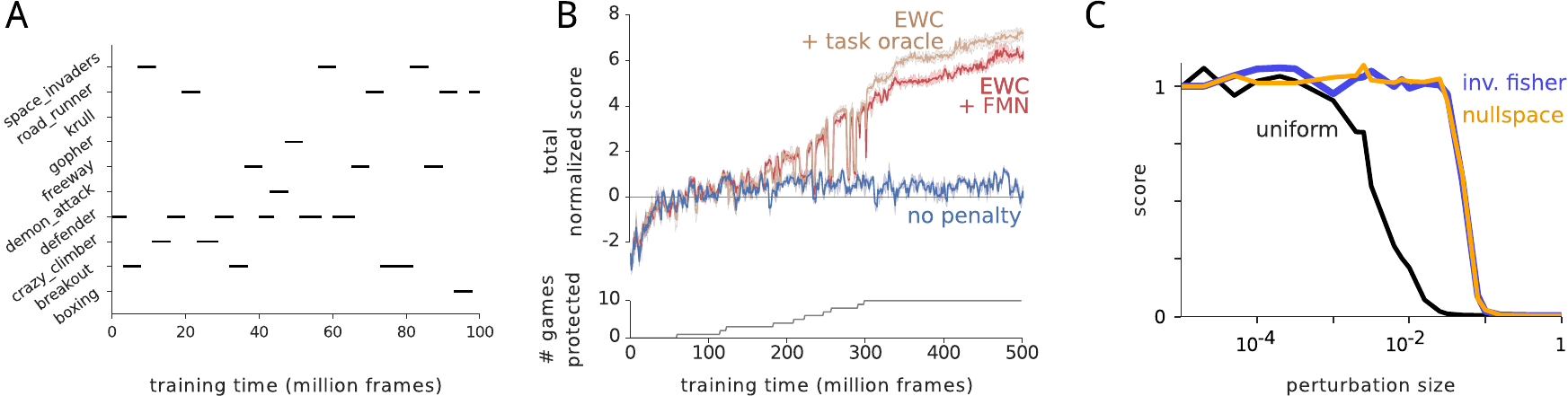}
	\end{center}
    \caption{\label{Atari10fig} Results on Atari task. A: Schedule of games. Black bars indicate the sequential training periods (segments) for each game. After each training segment, performance on all games is measured. The EWC constraint is only activated to protect an agent's performance on each game once the agent has experienced 20 million frames in that game. B: Total scores for each method across all games. Red curve denotes the network which infers the task labels using the Forget Me Not algorithm; brown curve is the network provided with the task labels. The EWC and SGD curves start diverging when games start being played again that have been protected by EWC. C: Sensitivity of a single-game DQN, trained on Breakout, to noise added to its weights. The performance on Breakout is shown as a function of the magnitude (standard deviation) of the weight perturbation. The weight perturbation is drawn from a zero mean Gaussian with covariance that is either uniform (black; i.e.\ targets all weights equally), the inverse Fisher ($(F + \lambda I)^{-1}$; blue; i.e.\ mimicking weight changes allowed by EWC), or uniform within the nullspace of the Fisher (orange; i.e.\ targets weights that the Fisher estimates that the network output is entirely invariant to). To evaluate the score, we ran the agent for ten full game episodes, drawing a new random weight perturbation for every timestep.
    }
\end{figure*}

We next tested whether elastic weight consolidation could support continual learning in the far more demanding reinforcement learning (RL) domain. In RL, agents dynamically interact with the environment in order to develop a policy that maximizes cumulative future reward. We asked whether Deep Q Networks (DQNs)---an architecture that has achieved impressive successes in such challenging RL settings \citep{mnih2015human}---could be harnessed with EWC to successfully support continual learning in the classic Atari 2600 task set \citep{bellemare2012arcade}. Specifically, each experiment consisted of ten games chosen randomly from those that are played at human level or above by DQN. At training time, the agent was exposed to experiences from each game for extended periods of time. The order of presentation of the games was randomized and allowed for returning to the same games several times. At regular intervals we would also test the agent's score on each of the ten games, without allowing the agent to train on them (Figure \ref{Atari10fig}A).

Notably, previous reinforcement learning approaches to continual learning have either relied on either adding capacity to the network \citep{ring1998child, rusu2016progressive} or on learning each task in separate networks, which are then used to train a single network that can play all games\citep{rusu2015policy, parisotto2015actor}. In contrast, the EWC approach presented here makes use of a single network with fixed resources (i.e.\ network capacity) and has minimal computational overhead.

In addition to using EWC to protect previously-acquired knowledge, we used the RL domain to address a broader set of requirements that are needed for successful continual learning systems: in particular, higher-level mechanisms are needed to infer which task is currently being performed, detect and incorporate novel tasks as they are encountered, and allow for rapid and flexible switching between tasks \citep{collins2013cognitive}. In the primate brain, the prefrontal cortex is widely viewed as supporting these capabilities by sustaining neural representations of task context that exert top-down gating influences on sensory processing, working memory, and action selection in lower-level regions \citep{oreilly2006making, mante2013context, miller2001integrative, doya2002multiple}.

Inspired by this evidence,  we used an agent very similar to that described in
\citep{van2015deep} with few differences: (a) a network with more parameters,
(b) a smaller transition table, (c) task-specific bias and gains at each layer,
(d) the full action set in Atari,
(e) a task-recognition model, and (e) the EWC penalty.
Full details of hyper-parameters are described in Appendix~\emph{app:atari}. Here we briefly describe the two most important modifications to the agent: the task-recognition module, and the implementation of the EWC penalty.

We treat the task context as the latent variable of a Hidden Markov Model. Each task is therefore associated to an underlying generative model of the observations. The main distinguishing feature of our approach is that we allow for the addition of new generative models if they explain recent data better than the existing pool of models by using a training procedure inspired by the forget me not process\citep{venessX} (see Appendix~\ref{app:atari}).

In order to apply EWC, we compute the Fisher information matrix at each task switch. For each task, a penalty is added with anchor point given by the current value of the parameters and with weights given by the Fisher information matrix times a scaling factor $\lambda$ which was optimized by hyperparameter search. We only added an EWC penalty to games which had experienced at least $20$ million frames.

We also allowed the DQN agents to maintain separate short-term memory buffers for each inferred task: these allow action values for each task to be learned off-policy using an experience replay mechanism \citep{mnih2015human}. As such, the overall system has memory on two time-scales: over short time-scales, the experience replay mechanism allows learning in DQN to be based on the interleaved and uncorrelated experiences \citep{mnih2015human}. At longer time scales, know-how across tasks is consolidated by using EWC. Finally, we allowed a small number of network parameters to be game-specific, rather than shared across games. {In particular, we allowed each layer of the network to have biases and per element multiplicative gains that were specific to each game.}

We compare the performance of agents which use EWC (red) with ones that do not (blue) over sets of ten games in Figure \ref{Atari10fig}. We measure the performance as the total human-normalized score across all ten games. We average across random seeds and over the choice of which ten games were played (see Appendix~\ref{app:atari}). We also clip the human-normalized score for each game to 1. Our measure of performance is therefore a number with a maximum of 10 (at least at human level on all games) where 0 means the agent is as good as a random agent.  If we rely on plain gradient descent methods as in \citep{mnih2015human}, the agent never learns to play more than one game and the harm inflicted by forgetting the old games means that the total human-normalized score remains below one. By using EWC, however, the agents do indeed learn to play multiple games. As a control, we also considered the benefit to the agent if we explicitly provided the agent with the true task label (Figure \ref{Atari10fig}B, brown), rather than relying on the learned task recognition through the FMN algorithm (red). The improvement here was only modest.

While augmenting the DQN agent with EWC allows it to learn many games in sequence without suffering from catastrophic forgetting, it does not reach the score that would have been obtained by training ten separate DQNs (see Figure 1 in Appendix~\ref{app:atari}).  One possible reason for this is that we consolidated weights for each game based on a tractable approximation of parameter uncertainty, the Fisher Information. We therefore sought to test the quality of our estimates empirically. To do so, we trained an agent on a single game, and measured how perturbing the network parameters affected the agent's score. Regardless of which game the agent was trained on, we observed the same patterns, shown in Figure \ref{Atari10fig}C. First, the agent was always more robust to parameter perturbations shaped by the inverse of the diagonal of the Fisher Information (blue), as opposed to uniform perturbations (black). This validates that the diagonal of the Fisher is a good estimate of how important a certain parameter is. {Within our approximation, perturbing in the nullspace should have no effect on performance at all on performance. Empirically, however, we observe that perturbing in this space (orange) has the same effect as perturbing in the inverse Fisher space. This suggests that we are over-confident about certain parameters being unimportant: }it is therefore likely that the chief limitation of the current implementation is that it under-estimates parameter uncertainty.

\section{Discussion}

We present a novel algorithm, elastic weight consolidation, that addresses the significant problem continual learning poses for neural networks. EWC allows knowledge of previous tasks to be protected during new learning, thereby avoiding catastrophic forgetting of old abilities. It does so by selectively decreasing the plasticity of weights, and thus has parallels with neurobiological models of synaptic consolidation. We implement EWC as a soft, quadratic constraint whereby each weight is pulled back towards its old values by an amount proportional to its importance for performance on previously-learnt tasks. To the extent that tasks share structure, networks trained with EWC reuse shared components of the network. We further show that EWC can be effectively combined with deep neural networks to support continual learning in challenging reinforcement learning scenarios, such as Atari 2600 games.

The EWC algorithm can be grounded in Bayesian approaches to learning. Formally, when there is a new task to be learnt, the network parameters are tempered by a prior which is the \emph{posterior} distribution on the parameters given data from previous task(s). This enables fast learning rates on parameters that are poorly constrained by the previous tasks, and slow learning rates for those which are crucial.

There has been previous work \citep{french2002using, eaton2013ella} using a quadratic penalty to approximate old parts of the dataset, but these applications have been limited to small models.  Specifically, \citep{french2002using} used random inputs to compute a quadratic approximation to the energy surface. Their approach is slow, as it requires re-computing the curvature at each sample. The ELLA algorithm described in \citep{eaton2013ella} requires computing and inverting matrices with a dimensionality equal to the number of parameters being optimized, therefore it has been mainly applied to linear and logistic regressions. In contrast, EWC has a run time which is linear in both the number of parameters and the number of training examples. We could only achieve this low computational complexity by making several simplifications, most notably by approximating the posterior distribution of the parameters on a task (i.e.\ the weight uncertainties) by a factorized Gaussian, and by computing its variance using a point-estimate of the parameters, via the diagonal of the Fisher Information matrix. Despite its low computational cost and empirical successes---even in the setting of challenging RL domains---our use of a point estimate of the posterior's variance (as in a Laplace approximation) does constitute a significant weakness (see Fig 4C). Our initial explorations suggest that one might improve on this local estimate by using Bayesian neural networks \citep{blundell2015weight}.

While this paper has primarily focused on building an algorithm out of neurobiological observations, it is also instructive to consider whether the algorithm's successes can feed back into our understanding of the brain. In particular, we see considerable parallels between EWC and two computational theories of synaptic plasticity.

In this respect, the perspective we offer here aligns with a recent proposal that each synapse not only stores its current weight, but also an implicit representation of its uncertainty about that weight \citep{aitchison2015synaptic}. This idea is grounded in observations that post-synaptic potentials are highly variable in amplitude (suggestive of sampling from the weight posterior during computation), and that those synapses which are more variable are more amenable to potentiation or depression (suggestive of updating the weight posterior). While we do not explore the computational benefits of sampling from a posterior here, our work aligns with the notion that weight uncertainty should inform learning rates. We take this one step further, to emphasize that consolidating the high precision weights enables continual learning over long time scales. {With EWC, three values have to be stored for each synapse: the weight itself, its variance and its mean. Interestingly, synapses in the brain also carry more than one piece of information. For example, the state of the short-term plasticity could carry information on the variance \citep{aitchison2015synaptic, pfister2010synapses}. The weight for the early phase of plasticity \citep{clopath2008tag} could encode the current synaptic strength, whereas the weight associated with the late-phase of plasticity or the consolidated phase could encode the mean weight.}

The ability to learn tasks in succession without forgetting is a core component of biological and artificial intelligence. In this work we show that an algorithm that supports continual learning---which takes inspiration from neurobiological models of synaptic consolidation---can be combined with deep neural networks to achieve successful performance in a range of challenging domains. In doing so, we demonstrate that current neurobiological theories concerning synaptic consolidation do indeed scale to large-scale learning systems. This provides prima facie evidence that these principles may be fundamental aspects of learning and memory in the brain.

\noindent
\textbf{Acknowledgements.} We would like to thank P. Dayan, D. Wierstra, S. Mohamed, Yee Whye Teh and K. Kavukcuoglu.




\nocite{veness2012context}
\nocite{dowson1982frechet}
\bibliographystyle{plainnat}
\bibliography{doc}

\begin{thebibliography}{42}
\providecommand{\natexlab}[1]{#1}
\providecommand{\url}[1]{\texttt{#1}}
\expandafter\ifx\csname urlstyle\endcsname\relax
  \providecommand{\doi}[1]{doi: #1}\else
  \providecommand{\doi}{doi: \begingroup \urlstyle{rm}\Url}\fi

\bibitem[Aitchison and Latham(2015)]{aitchison2015synaptic}
Laurence Aitchison and Peter~E Latham.
\newblock Synaptic sampling: A connection between psp variability and
  uncertainty explains neurophysiological observations.
\newblock \emph{arXiv preprint arXiv:1505.04544}, 2015.

\bibitem[Bellemare et~al.(2013)Bellemare, Naddaf, Veness, and
  Bowling]{bellemare2012arcade}
Marc~G Bellemare, Yavar Naddaf, Joel Veness, and Michael Bowling.
\newblock The arcade learning environment: An evaluation platform for general
  agents.
\newblock \emph{Journal of Artificial Intelligence Research}, 47:\penalty0
  253--279, 2013.

\bibitem[Benna and Fusi(2016)]{benna2015computational}
Marcus~K Benna and Stefano Fusi.
\newblock Computational principles of synaptic memory consolidation.
\newblock \emph{Nature neuroscience}, 2016.

\bibitem[Blundell et~al.(2015)Blundell, Cornebise, Kavukcuoglu, and
  Wierstra]{blundell2015weight}
Charles Blundell, Julien Cornebise, Koray Kavukcuoglu, and Daan Wierstra.
\newblock Weight uncertainty in neural network.
\newblock In \emph{Proceedings of The 32nd International Conference on Machine
  Learning}, pages 1613--1622, 2015.

\bibitem[Cichon and Gan(2015)]{cichon2015branch}
Joseph Cichon and Wen-Biao Gan.
\newblock Branch-specific dendritic ca2+ spikes cause persistent synaptic
  plasticity.
\newblock \emph{Nature}, 520\penalty0 (7546):\penalty0 180--185, 2015.

\bibitem[Clopath et~al.(2008)Clopath, Ziegler, Vasilaki, B{\"u}sing, and
  Gerstner]{clopath2008tag}
Claudia Clopath, Lorric Ziegler, Eleni Vasilaki, Lars B{\"u}sing, and Wulfram
  Gerstner.
\newblock Tag-trigger-consolidation: a model of early and late
  long-term-potentiation and depression.
\newblock \emph{PLoS Comput Biol}, 4\penalty0 (12):\penalty0 e1000248, 2008.

\bibitem[Collins and Frank(2013)]{collins2013cognitive}
Anne~GE Collins and Michael~J Frank.
\newblock Cognitive control over learning: creating, clustering, and
  generalizing task-set structure.
\newblock \emph{Psychological review}, 120\penalty0 (1):\penalty0 190, 2013.

\bibitem[Dowson and Landau(1982)]{dowson1982frechet}
DC~Dowson and BV~Landau.
\newblock The fr{\'e}chet distance between multivariate normal distributions.
\newblock \emph{Journal of multivariate analysis}, 12\penalty0 (3):\penalty0
  450--455, 1982.

\bibitem[Doya et~al.(2002)Doya, Samejima, Katagiri, and
  Kawato]{doya2002multiple}
Kenji Doya, Kazuyuki Samejima, Ken-ichi Katagiri, and Mitsuo Kawato.
\newblock Multiple model-based reinforcement learning.
\newblock \emph{Neural computation}, 14\penalty0 (6):\penalty0 1347--1369,
  2002.

\bibitem[Eaton and Ruvolo(2013)]{eaton2013ella}
Eric Eaton and Paul~L Ruvolo.
\newblock Ella: An efficient lifelong learning algorithm.
\newblock In \emph{International Conference on Machine Learning}, pages
  507--515, 2013.

\bibitem[Eskin et~al.(2004)Eskin, Smola, and Vishwanathan]{smola2003laplace}
Eleazar Eskin, Alex~J. Smola, and S.v.n. Vishwanathan.
\newblock Laplace propagation.
\newblock In \emph{Advances in Neural Information Processing Systems 16}, pages
  441--448. MIT Press, 2004.
\newblock URL \url{http://papers.nips.cc/paper/2444-laplace-propagation.pdf}.

\bibitem[French(1999)]{french1999catastrophic}
Robert~M French.
\newblock Catastrophic forgetting in connectionist networks.
\newblock \emph{Trends in cognitive sciences}, 3\penalty0 (4):\penalty0
  128--135, 1999.

\bibitem[French and Chater(2002)]{french2002using}
Robert~M French and Nick Chater.
\newblock Using noise to compute error surfaces in connectionist networks: a
  novel means of reducing catastrophic forgetting.
\newblock \emph{Neural computation}, 14\penalty0 (7):\penalty0 1755--1769,
  2002.

\bibitem[Fusi et~al.(2005)Fusi, Drew, and Abbott]{fusi2005cascade}
Stefano Fusi, Patrick~J Drew, and LF~Abbott.
\newblock Cascade models of synaptically stored memories.
\newblock \emph{Neuron}, 45\penalty0 (4):\penalty0 599--611, 2005.

\bibitem[Goodfellow et~al.(2014)Goodfellow, Mirza, Xiao, Courville, and
  Bengio]{goodfellow2015empirical}
Ian~J Goodfellow, Mehdi Mirza, Da~Xiao, Aaron Courville, and Yoshua Bengio.
\newblock An empirical investigation of catastrophic forgeting in
  gradient-based neural networks.
\newblock \emph{Int'l Conf. on Learning Representations (ICLR)}, 2014.

\bibitem[Hayashi-Takagi et~al.(2015)Hayashi-Takagi, Yagishita, Nakamura,
  Shirai, Wu, Loshbaugh, Kuhlman, Hahn, and Kasai]{hayashi2015labelling}
Akiko Hayashi-Takagi, Sho Yagishita, Mayumi Nakamura, Fukutoshi Shirai, Yi~I.
  Wu, Amanda~L. Loshbaugh, Brian Kuhlman, Klaus~M. Hahn, and Haruo Kasai.
\newblock Labelling and optical erasure of synaptic memory traces in the motor
  cortex.
\newblock \emph{Nature}, 525\penalty0 (7569):\penalty0 333--338, 09 2015.
\newblock URL \url{http://dx.doi.org/10.1038/nature15257}.

\bibitem[Kieran et~al.(2016)Kieran, Veness, Bowling, Kirkpatrick, Koop, and
  Hassabis]{venessX}
Milan Kieran, Joel Veness, Michael Bowling, James Kirkpatrick, Anna Koop, and
  Demis Hassabis.
\newblock The forget me not process.
\newblock In \emph{Advances in Neural Information Processing Systems 26}, page
  accepted for publication, 2016.

\bibitem[Krizhevsky et~al.(2012)Krizhevsky, Sutskever, and
  Hinton]{krizhevsky2012imagenet}
Alex Krizhevsky, Ilya Sutskever, and Geoffrey~E Hinton.
\newblock Imagenet classification with deep convolutional neural networks.
\newblock In \emph{NIPS}, pages 1097--1105, 2012.

\bibitem[LeCun et~al.(1998)LeCun, Cortes, and Burges]{lecun1998mnist}
Yann LeCun, Corinna Cortes, and Christopher~JC Burges.
\newblock The mnist database of handwritten digits, 1998.

\bibitem[LeCun et~al.(2015)LeCun, Bengio, and Hinton]{lecun2015deep}
Yann LeCun, Yoshua Bengio, and Geoffrey Hinton.
\newblock Deep learning.
\newblock \emph{Nature}, 521\penalty0 (7553):\penalty0 436--444, 2015.

\bibitem[Legg and Hutter(2007)]{legg2007universal}
Shane Legg and Marcus Hutter.
\newblock Universal intelligence: A definition of machine intelligence.
\newblock \emph{Minds and Machines}, 17\penalty0 (4):\penalty0 391--444, 2007.

\bibitem[MacKay(1992)]{mackay1992practical}
David~JC MacKay.
\newblock A practical bayesian framework for backpropagation networks.
\newblock \emph{Neural computation}, 4\penalty0 (3):\penalty0 448--472, 1992.

\bibitem[Mante et~al.(2013)Mante, Sussillo, Shenoy, and
  Newsome]{mante2013context}
Valerio Mante, David Sussillo, Krishna~V Shenoy, and William~T Newsome.
\newblock Context-dependent computation by recurrent dynamics in prefrontal
  cortex.
\newblock \emph{Nature}, 503\penalty0 (7474):\penalty0 78--84, 2013.

\bibitem[McClelland et~al.(1995)McClelland, McNaughton, and
  O'Reilly]{mcclelland1995}
James~L McClelland, Bruce~L McNaughton, and Randall~C O'Reilly.
\newblock Why there are complementary learning systems in the hippocampus and
  neocortex: insights from the successes and failures of connectionist models
  of learning and memory.
\newblock \emph{Psychological review}, 102\penalty0 (3):\penalty0 419, 1995.

\bibitem[McCloskey and Cohen(1989)]{mccloskey1989catastrophic}
Michael McCloskey and Neal~J Cohen.
\newblock Catastrophic interference in connectionist networks: The sequential
  learning problem.
\newblock \emph{The psychology of learning and motivation}, 24\penalty0
  (109-165):\penalty0 92, 1989.

\bibitem[Miller and Cohen(2001)]{miller2001integrative}
Earl~K Miller and Jonathan~D Cohen.
\newblock An integrative theory of prefrontal cortex function.
\newblock \emph{Annual review of neuroscience}, 24\penalty0 (1):\penalty0
  167--202, 2001.

\bibitem[Mnih et~al.(2015)Mnih, Kavukcuoglu, Silver, Rusu, Veness, Bellemare,
  Graves, Riedmiller, Fidjeland, Ostrovski, et~al.]{mnih2015human}
Volodymyr Mnih, Koray Kavukcuoglu, David Silver, Andrei~A Rusu, Joel Veness,
  Marc~G Bellemare, Alex Graves, Martin Riedmiller, Andreas~K Fidjeland, Georg
  Ostrovski, et~al.
\newblock Human-level control through deep reinforcement learning.
\newblock \emph{Nature}, 518\penalty0 (7540):\penalty0 529--533, 2015.

\bibitem[Nielsen(1989)]{hechtnielsen:theory}
Robert~H. Nielsen.
\newblock Theory of the backpropagation neural network.
\newblock In \emph{Proceedings of the International Joint Conference on Neural
  Networks}, volume~I, pages 593--605. Piscataway, NJ: IEEE, 1989.

\bibitem[O'Reilly and Frank(2006)]{oreilly2006making}
Randall~C O'Reilly and Michael~J Frank.
\newblock Making working memory work: a computational model of learning in the
  prefrontal cortex and basal ganglia.
\newblock \emph{Neural computation}, 18\penalty0 (2):\penalty0 283--328, 2006.

\bibitem[Parisotto et~al.(2015)Parisotto, Ba, and
  Salakhutdinov]{parisotto2015actor}
Emilio Parisotto, Jimmy~Lei Ba, and Ruslan Salakhutdinov.
\newblock Actor-mimic: Deep multitask and transfer reinforcement learning.
\newblock \emph{arXiv preprint arXiv:1511.06342}, 2015.

\bibitem[Pascanu and Bengio(2013)]{pascanu2013revisiting}
Razvan Pascanu and Yoshua Bengio.
\newblock Revisiting natural gradient for deep networks.
\newblock \emph{arXiv preprint arXiv:1301.3584}, 2013.

\bibitem[Pfister et~al.(2010)Pfister, Dayan, and Lengyel]{pfister2010synapses}
Jean-Pascal Pfister, Peter Dayan, and M{\'a}t{\'e} Lengyel.
\newblock Synapses with short-term plasticity are optimal estimators of
  presynaptic membrane potentials.
\newblock \emph{Nature neuroscience}, 13\penalty0 (10):\penalty0 1271--1275,
  2010.

\bibitem[Ratcliff(1990)]{ratcliff1990connectionist}
Roger Ratcliff.
\newblock Connectionist models of recognition memory: constraints imposed by
  learning and forgetting functions.
\newblock \emph{Psychological review}, 97\penalty0 (2):\penalty0 285, 1990.

\bibitem[Ring(1998)]{ring1998child}
Mark~B Ring.
\newblock Child: A first step towards continual learning.
\newblock In \emph{Learning to learn}, pages 261--292. Springer, 1998.

\bibitem[Rusu et~al.(2015)Rusu, Colmenarejo, Gulcehre, Desjardins, Kirkpatrick,
  Pascanu, Mnih, Kavukcuoglu, and Hadsell]{rusu2015policy}
Andrei~A Rusu, Sergio~Gomez Colmenarejo, Caglar Gulcehre, Guillaume Desjardins,
  James Kirkpatrick, Razvan Pascanu, Volodymyr Mnih, Koray Kavukcuoglu, and
  Raia Hadsell.
\newblock Policy distillation.
\newblock \emph{arXiv preprint arXiv:1511.06295}, 2015.

\bibitem[Rusu et~al.(2016)Rusu, Rabinowitz, Desjardins, Soyer, Kirkpatrick,
  Kavukcuoglu, Pascanu, and Hadsell]{rusu2016progressive}
Andrei~A Rusu, Neil~C Rabinowitz, Guillaume Desjardins, Hubert Soyer, James
  Kirkpatrick, Koray Kavukcuoglu, Razvan Pascanu, and Raia Hadsell.
\newblock Progressive neural networks.
\newblock \emph{arXiv preprint arXiv:1606.04671}, 2016.

\bibitem[Srivastava et~al.(2013)Srivastava, Masci, Kazerounian, Gomez, and
  Schmidhuber]{srivastava2013compete}
Rupesh~K Srivastava, Jonathan Masci, Sohrob Kazerounian, Faustino Gomez, and
  Juergen Schmidhuber.
\newblock Compete to compute.
\newblock In \emph{Advances in Neural Information Processing Systems 26}, pages
  2310--2318. Curran Associates, Inc., 2013.
\newblock URL \url{http://papers.nips.cc/paper/5059-compete-to-compute.pdf}.

\bibitem[Sussmann(1992)]{Sussmann1992UniquenessOT}
H{\'e}ctor~J. Sussmann.
\newblock Uniqueness of the weights for minimal feedforward nets with a given
  input-output map.
\newblock \emph{Neural Networks}, 5:\penalty0 589--593, 1992.

\bibitem[van Hasselt et~al.(2016)van Hasselt, Guez, and Silver]{van2015deep}
Hado van Hasselt, Arthur Guez, and David Silver.
\newblock Deep reinforcement learning with double q-learning.
\newblock \emph{Proceedings of the Thirtieth AAAI Conference on Artificial
  Intelligence}, pages 2094--2100, 2016.

\bibitem[Veness et~al.(2012)Veness, Ng, Hutter, and Bowling]{veness2012context}
Joel Veness, Kee~Siong Ng, Marcus Hutter, and Michael Bowling.
\newblock Context tree switching.
\newblock In \emph{2012 Data compression conference.}, pages 327--336. IEEE,
  2012.

\bibitem[Yang et~al.(2009)Yang, Pan, and Gan]{yang2009stably}
Guang Yang, Feng Pan, and Wen-Biao Gan.
\newblock Stably maintained dendritic spines are associated with lifelong
  memories.
\newblock \emph{Nature}, 462\penalty0 (7275):\penalty0 920--924, 2009.

\bibitem[Yang et~al.(2014)Yang, Lai, Cichon, Ma, Li, and Gan]{yang2014sleep}
Guang Yang, Cora Sau~Wan Lai, Joseph Cichon, Lei Ma, Wei Li, and Wen-Biao Gan.
\newblock Sleep promotes branch-specific formation of dendritic spines after
  learning.
\newblock \emph{Science}, 344\penalty0 (6188):\penalty0 1173--1178, 2014.

\end{thebibliography}

\newpage

\section{Appendix}

\subsection{MNIST experiments}
\label{app:mnist}
We carried out all MNIST experiments with fully-connected networks with rectified linear units.  In order to replicate the results of \citep{goodfellow2015empirical}, we compared to results obtained using dropout regularization. As suggested in \citep{goodfellow2015empirical}, we applied dropout with a probability of 0.2 to the input and of 0.5 to the other hidden layers. In order to give SGD with dropout the best possible chance, we also used early stopping. Early stopping was implemented by computing the test error on the validation set for all pixel permutations seen to date. Here, if the validation error was observed to increase for more than five subsequent steps, we terminated this training segment and proceeded to the next dataset; at this point, we reset the network weights to the values that had the lowest average validation error on all previous datasets.  Table \ref{mnist_hyper} shows a list of all hyperparameters used to produce the three graphs in Figure 3 of the main text. Where a range is present, the parameter was randomly varied and the reported results were obtained using the best hyperparameter setting. When random hyperparameter search was used, 50 combinations of parameters were attempted for each number experiment.

\begin{table}
  \caption{\label{mnist_hyper} Hyperparameters for each of the MNIST figures}
  \centering
  \begin{tabular}{l ccc}
    \hline \hline
    Hyperparameter   & \multicolumn{3}{c}{Reference figure} \\
                     & 3A         & 3B              &  3C \\
    learning rate          &  $10^{-3}$    & $10^{-5}$-$10^{-3}$   & $10^{-3}$ \\
    dropout                &  no        & yes             & no \\
    early stopping         &  no        & yes             & no \\
    n. hidden layers       &  2         & 2               & 6  \\
    width hidden layers    &  400       & 400-2000        & 100 \\
    epochs / dataset       &  20        & 100             & 100
  \end{tabular}
\end{table}

\subsection{Atari experiments}
\label{app:atari}
The agent architecture used is almost identical to that used in \citep{van2015deep}.
In this section we provide details on all the parameters used.

Images are preprocessed in the same way as in \citep{mnih2015human}, namely
the 210x160 images from the Atari emulator are downsampled to 84x84 using
bilinear interpolation. We then convert the RGB images to YUV and use the
grayscale channel alone. The state used by the agent consists of the four
latest downsampled, grayscale observations concatenated together.

The network structure used is similar to the one from \citep{mnih2015human}, namely three convolutional layers followed by a fully connected layer.  The first convolution had kernel size 8, stride 4 and 32 filters.  The second convolution had kernel size 4, stride 2 and 64 filters.  The final convolution had kernels size 3, stride 1 and 128 filters.  The fully connected layer had 1024 units. Note that this network has approximately four times as many parameters as the standard network, due to having twice as many fully connected units and twice as many filters in the final convolution. The other departure from the standard network is that each layer was allowed to have task-specific gains and biases.  For each layer, the transformation $x \rightarrow y$ computed by the network is therefore:
\begin{equation}
\label{gainsAndBiases}
y_i = \left( \sum_j W_{ij} x_j + b^{c}_i \right) g^{c}_i
\end{equation}
where the biases $b$ and the gains $g$.
The network weights and biases where initialized by setting them randomly with a uniform number between $-\sigma$ and $\sigma$, with $\sigma$ set to the square root of the incoming hidden units (for a linear layer) or set to the area of the kernel times the number of incoming filters (for convolutional layers). Biases and gains were initialized to 0 and 1 respectively.

We used an an $\epsilon$-greedy exploration policy, where the probability of selecting random action, $\epsilon$, decayed with training time. We kept a different timer for each of the tasks. We set $\epsilon = 1$ for $5 \times 10^4$ time steps, and then decayed this linearly to a value of 0.01 for the next $10^6$.

We trained the networks with the Double Q-learning algorithm \citep{van2015deep}. A training step is carried out on a minibatch of 32 experiences every four steps. The target network is updated every $3 \times 10^4$ time steps. We trained with RMSProp, with a momentum of $0.$, a decay of $0.95$, a learning rate of $2.5 \times 10^{-4}$, and a maximum learning rate of $ 2.5 \times 10^{-3}$.

Other hyperparameters that we changed from the reference implementation were: 1) using a smaller replay buffer ($5 \times 10^5$ past experiences), and 2) a scaling factor for the EWC penalty of $400$. Another subtle difference is that we used the full action set in the Atari emulator. In fact, although many games only support a small subset of the 18 possible actions, in order to have a unified network structure for all games we used 18 actions in each game.

\begin{figure}
\begin{center}
\includegraphics[width=\textwidth,trim={2cm 2cm 2cm 2cm},clip]{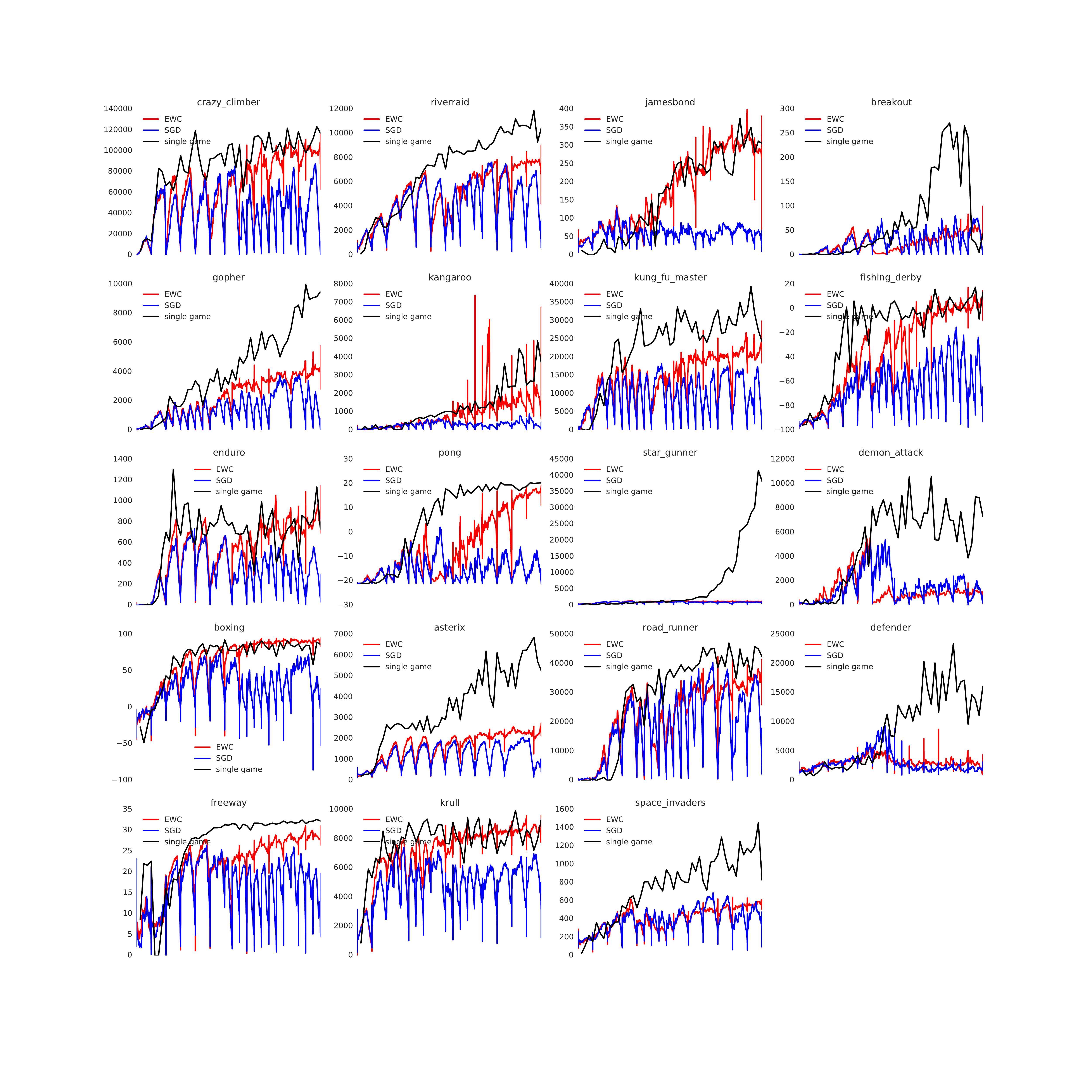}
\end{center}
\caption{\label{Atari10-byGame} Score in the individual games as a function of
    steps played in that game. The black baseline curves show learning on
    individual games alone.}
\end{figure}

We randomly chose the 10 games for each experiment from a pool of 19 Atari games for which the standalone DQN could reach human-level performance in $50 \times 10^6$ frames. The scores for each of these games for the baseline algorithm, for EWC and for plain SGD training, as a function of the number steps played \emph{in that game} are shown in Figure \ref{Atari10-byGame}. In order to get an averaged performance, we chose 10 sets of 10 games, and ran 4 different random seeds for each set. 

The most significant departure from the published models is the automatic determination of the task.  We model each task by a generative model of the environment. In this work, for simplicity, we only model the current observation. The current task is modelled as a categorical context $c$ which is treated as the hidden variable in an Hidden Markov Model that explain observations. In such a model the probabilty of being in a particular context $c$ at time $t$ evolves according to:
\begin{align*}
p(c, t+1) &= \sum_{c'} p(c', t) \Gamma(c, c')  \\
\Gamma(c, c') &= \delta(c, c') (1-\alpha) + (1-\delta(c, c')) \alpha
\end{align*}

where $\delta$ is the Kronecker delta function and $\alpha$ is the probability of switching context. The task context then conditions a generative model predicting the observation probability $p(o|c, t)$.  Given such generative models, the probability of being in a task set at time $ t $ can be inferred by the observations seen so far as:
\begin{equation*}
p(c\ | o_1 ... o_t) \propto \sum_{c'} \Gamma(c, c')\  p(c', t-1)  p(o|c, t)
\end{equation*}
The maximal probability context is then taken to be the current task label.

In our implementation, the generative models consist of factored multinomial distributions explaining the probability of the state of each pixel in the observation space.  The model is a parametrized Dirichlet distribution, which summarizes the data seen so far using Bayesian updates.  In order to encourage each model to specialize, we train the models as follows. We partition time into windows of a particular width $ W $. During each window, all the Dirichlet priors are updated with the evidence seen so far. At the end of the window, the model best corresponding to the current task set is selected.  Since this model was the most useful to explain the current data, it keeps its prior, while all other priors are reverted to their state at the beginning of the time window. We ensure that one hold-out uniform (\emph{i.e.} uninitialized) Dirichlet-multinomial is always available. Whenever the hold-out model is selected a new generative model is created and a new task context is therefore created. This model is Bayesian, in the sense that data is used to maintain beliefs over priors on the generative models, and is non-parametric, in the sense that the model can grow in function of the observed data. It can be seen as an implementation of the flat forget me not algorithm described in \citep{venessX}. The parameter $\alpha$ is not learnt. Instead we use the result from \citep{veness2012context} where it is shown that a time decaying switch rate $ \alpha = 1/t$ guarantees good worst case asymptotic perfmance provided the number of tasks grows as $o\left(\frac{n}{\log n}\right)$.

Table \ref{atari_hyper} summarizes all hyper-parameters used for the Atari experiments. Except for the parameters pertaining the EWC algorithm (Fisher multiplier, num. samples Fisher, EWC start) or pertaining the task recognition models (model update period, model downscaling and size window), all the parameters values are the same as from \citep{van2015deep} and have not been tuned for these experiments. 
\begin{table*}
  \caption{\label{atari_hyper} Hyperparameters for each of the MNIST figures}
  \centering
  \begin{tabular}{l cl}
    \hline \hline
    Hyperparameter         & value   & brief description \\
    \hline
    action repeat          & 4       & Repeat the same action for four frames. Each agent step will\\
    & & occur every fourth frame. \\
    discount factor        & 0.99    & Discount factor used in the Q-learning algorithm. \\
    no-op max              & 30      & Maximum number of do nothing operations carried out at the \\ 
    & & beginning of each training episode to provide a  varied training set.\\
    max. reward            & 1       & Rewards are clipped to 1. \\
    scaled input           & 84x84   & Input images are scaled to 84x84 with bilinear interpolation. \\
    optimization algorithm & RMSprop & Optimization algorithm used. \\
    learning rate          & 0.00025 & The learning rate in RMSprop. \\
    max. learning rate     & 0.0025  & The maximum learning rate that RMSprop will apply.\\
    momentum               & 0.      & The momentum used in RMSprop.\\
    decay                  & 0.95    & The decay used in RMSProp. \\
    clip$\delta$           & 1.      & Each gradient from Q-learning is clipped to $\pm$ 1. \\
    max. norm              & 50.     & After clipping, if the norm of the gradient is greater \\
    & & than 50., the gradient is renormalized to 50. \\
    history length         & 4       & The four most recently experienced frames are taken to \\
    & & form a state for Q-learning \\
    minibatch size         & 32      & The number of elements taken from the replay buffer to form\\
    & & a mini-batch training example.\\
    replay period          & 4       & A mini-batch is loaded from the replay buffer every 4 \\
    & & steps (16 frames including action repeat). \\ 
    memory size            & 50000   & The replay memory stores the last fify thousand \\
    & & transitions experienced.\\
    target update period   & 7500    & The target network in Q-learning is updated to the policy\\
    & & network every 7500 step. \\
    min. history           & 50000   & The agent will only start learning after fifty thousand\\
    & & transitions have been stored into memory.\\
    initial exploration    & 1.      & The value of the initial exploration rate. \\
    exploration decay start & 50000  & The exploration rate will start decaying after fifty \\
    & & thousand frames.\\
    exploration decay end   & 1050000 & The exploration rate will decay over one million frames.\\
    final exploration      & 0.01    & The value of the final exploration rate. \\
    model update period    & 4      & The Dirichlet model is updated every fourth step.\\
    model downscaling      & 2      & The Dirichlet model is downscaled by a factor of 2, that \\
    & & is an image of size 42x42 is being modelled.\\
    size window            & 4      & The size of the window for the task recognition model learning.\\
    num. samples Fisher   & 100     & Whenever the diagonal of the Fisher is recomputed for a \\
    & & task, one hundred mini-batches are drawn from the replay buffer. \\
    Fisher multiplier      & 400    & The Fisher is scaled by this number to form the EWC penalty.\\
    start EWC              & 20E6   & The EWC penalty is only applied after 5 million steps \\
    & & (20 million frames).
  \end{tabular}
\end{table*}

\subsection{Fisher overlap}
\label{app:fisher}

To assess whether different tasks solved in the same network use similar sets of weights (Figure 3C in the mains text), we measured the degree of overlap between the two tasks' Fisher matrices. Precisely, we computed two tasks' Fishers, $F_1$ and $F_2$, normalized these to each have unit trace, $\hat{F}_1$ and $\hat{F}_2$, then computed their Fr\'{e}chet distance, a metric on the space of positive-semidefinite matrices \citep{dowson1982frechet}:
\begin{align*}
  d^2(\hat{F}_1, \hat{F}_2)
    &= \frac 1 2 \mathrm{tr} \left( \hat{F}_1 + \hat{F}_2 - 2 (\hat{F}_1 \hat{F}_2)^{1/2} \right) \\
    &= \frac 1 2 || \hat{F}_1^{1/2} - \hat{F}_2^{1/2} ||_F
\end{align*}
which is bounded between zero and one. We then define the overlap as $1 - d^2$, with a value of zero indicating that the two tasks depend on non-overlapping sets of weights, and a value of one indicating that $F_1 = \alpha F_2$ for some $\alpha > 0$.

\end{document}